\documentclass[lettersize,journal]{IEEEtran}
\usepackage{amsmath,amsfonts}
\usepackage{array}
\usepackage[caption=false,font=normalsize,labelfont=sf,textfont=sf]{subfig}
\usepackage{textcomp}
\usepackage{stfloats}
\usepackage{url}
\usepackage{verbatim}
\usepackage{graphicx}
\usepackage{cite}
\usepackage{booktabs}
\usepackage{multirow}
\usepackage{xcolor}
\usepackage{algorithmic} 
\usepackage{algorithm}
\begin{document}

\title{No Prompt, No Leaks: A Robust Generative Steganography Framework via Prompt-Free Diffusion}

\author{Jingwen Cai, Fen Xiao, Shuhua Deng, Xieping Gao, ~\IEEEmembership{Member,~IEEE,}
\thanks{Jingwen Cai, Fen Xiao and Shuhua Deng are with the MOE Key Laboratory of Intelligent Computing and Information Processing, Xiangtan University, Xiangtan 411105, 
China (E-mail: Jenniewer\_cici@163.com; xiaof@xtu.edu.cn; shuhuadeng@xtu.edu.cn). (Corresponding authors: Fen Xiao and Shuhua Deng)}
\thanks{Xieping Gao is with the College of Information Science and Engineering, Hunan Normal University, Changsha 410006, China (E-mail: xpgao@hunnu.edu.cn).} 
\thanks{Manuscript received May 15, 2026.}}

\markboth{Journal of \LaTeX\ Class Files,~Vol.~14, No.~8, August~2021}%
{Shell \MakeLowercase{\textit{et al.}}: A Sample Article Using IEEEtran.cls for IEEE Journals}


\maketitle

\begin{abstract}
Generative image steganography synthesizes stego images directly from secret information to achieve inherent security advantages. Latent Diffusion Models (LDMs) have recently emerged as a fundamental image steganography framework that modulates secret latent representations with text prompts. Limited by the inflexibility of text prompts, these methods still struggle to generate high-quality stego images and accurately recover secret images. In this work, we propose a prompt-free diffusion image steganography framework that integrates style semantic priors to control more robust and reliable stego image generation. Specifically, a Cascaded Affine Coupling Module (CACM) establishes a bijective, deterministic mapping between a secret image and its latent representation. Then, style semantics are integrated into the diffusion process to control latent representation and ensure visual imperceptibility in the generated stego images. To mitigate trajectory deviations stemming from the unconditioned reverse process, a predictor-corrector mechanism is introduced to iteratively refine the generation trajectory via feedback from the current and predicted next states. Extensive experimental results show that the proposed method achieves competitive performance compared to state-of-the-art methods in terms of security, secret image reconstruction accuracy and controllability. 
\end{abstract}


\section{Introduction}
Image steganography conceals secret information within a cover image to enable secure and imperceptible transmission. It has been widely deployed in various security-critical scenarios, including copyright protection\cite{lv2025dopsteg}, digital watermarking\cite{zeng2025dualmodal}, and secure data communication\cite{ni2025face}. Existing methods inherently involve a three-way trade-off among payload capacity, security against steganalysis, and robustness against channel distortions and malicious attacks. Traditional image steganography typically embeds secret data directly into a cover image by minimizing a predefined distortion function. However, this embedding process inevitably introduces artifacts into the stego images that can be easily detected by advanced steganalysis tools, thereby severely limiting the security of the steganographic system~\cite{wei2022generative}.

Generative steganography leverages significant advances in generative models to directly synthesize the stego image from secret information without relying on the cover image~\cite{qin2019coverless}. This property theoretically enhances the undetectability of the stego images. Early studies employed generative adversarial networks (GANs)~\cite{goodfellow2014generative}, and flow-based models~\cite{kingma2018glow}, which mapped secret data to noise vectors~\cite{yu2021improved,wei2022flow,liu2022image, 9739773}. These methods can maintain high fidelity in recovered images. They suffer from insufficient robustness to cope with noise in real-world scenarios. To address this limitation, recent studies have introduced diffusion models into generative steganography~\cite{yu2023cross}. Latent Diffusion Models (LDMs) ~\cite{rombach2022high} have attracted significant attention due to their inherent noise robustness, powerful generative capabilities, and computational efficiency. In this paradigm, the secret image is embedded into the latent representation, in which both the visual content of the stego image and the secret recovery are guided by user-provided text prompts~\cite{xu2025diffusion, chen2025robust, jiang2026image,yang2024diffstega}.

\begin{figure}
  \centering
  \includegraphics[width=1\linewidth]{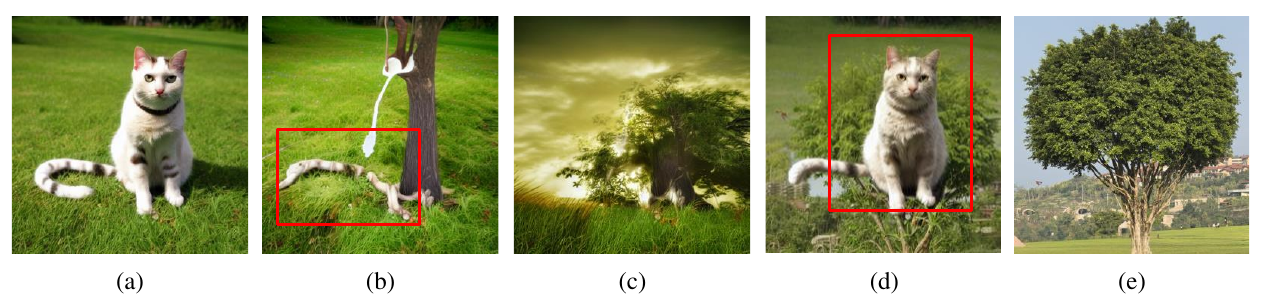}
  \caption{Impact of prompt type on stego image visual quality given the same secret image and semantic content. (a) Secret image. 
  (b) Coarse-grained text prompt (a tree). (c) Fine-grained text prompt (A massive ancient oak tree with deeply textured bark, complex branching structure, dense overlapping foliage, standing in a vast meadow at sunset). (d) Image prompt. (e) Reference image. }\label{fig11}
\end{figure}

Although text prompts offer enhanced controllability over the stego image, the semantic gap makes it difficult to ensure visual quality. Describing complex layouts, poses, shapes, and forms remains highly challenging and thus often leads to degradation in the generated stego image~\cite{zhang2023adding}. As shown in Fig.~\ref{fig11}, three stego images generated by CRoSS~\cite{yu2023cross} for the same secret image and share semantic content (“tree”), yet their visual quality differs dramatically depending on the type of text prompt. As shown in Fig.~\ref{fig11}(b), coarse-grained prompt "a tree" fails to sufficiently guide the generation process, which introduces obvious artifacts into the resulting stego image. While finer-grained prompts (e.g.a sentence), shows in Fig.~\ref{fig11}(c), can improve stego image quality to some extent, better aligning the generated images with user intent still requires numerous trial‑and‑error prompt edits. This process is not only often impractical but also inevitably increases the generation cost~\cite{zhang2023adding, xu2024prompt}. To reduce manual prompt engineering, we also follow the prior work~\cite{ye2023ipadapter} by using a reference image for structural guidance, as shown in Fig.~\ref{fig11}(e). However, the resulting stego image shown in Fig.~\ref{fig11}(d) leaks content both from the reference image and the secret image, which increases the risk of being detected by a steganalyzer.

Beyond the above security risks, the transmission of text prompts introduces additional reliability issues. Since the diffusion model requires the prompt as a conditioning signal for secret recovery, the prompt must be transmitted alongside the stego image. However, as a semantic key transmitted over practical channels, it is vulnerable to noise and tampering attacks~\cite{zhang2025provably}. Given the high sensitivity of diffusion models, even minor perturbations to the prompt can severely disrupt the recovery process~\cite{chen2025robust}.
This disruption is further aggravated by iterative accumulation, which gradually drives the recovery trajectory away from its intended path~\cite{Mokady2023CVPR, Wallace2023CVPR} and ultimately leads to failed secret reconstruction. Moreover, relying on a reference image as the recovery condition introduces lossy transmission~\cite{hu2024establishing}, which further exacerbates the system's unreliability in real‑world scenarios.

To address the above security and reliability challenges, we propose a prompt-free diffusion steganographic framework for robust image steganography.  We first discuss replacing text prompts with semantic priors to guide realistic stego image generation. These priors, extracted from reference datasets via a Vision Transformer (ViT)\cite{dosovitskiy2021image}, are used to avoid content-specific overfitting in the stego process, thereby improving both generation security and visual quality. Although this design eliminates the transmission risk associated with text prompts, the absence of explicit semantic guidance inevitably increases the uncertainty of the reverse trajectory. An unconditional way is introduced to stabilize the recovery process in the prompt-free setting. This framework offers several advantages, including user-controllable stego generation and enhanced robustness against real-world transmission distortions. 
The main contributions of this work are as follows.
\begin{enumerate}
\item We introduce a prompt-free stego image generation scheme with semantic priors. A cascaded affine coupling module (CACM) is designed to establish a deterministic mapping between the secret image and its latent representation. Subsequently, semantic priors obtained from reference datasets via ViT are incorporated into this representation to guide authentic stego image generation. This approach substantially enhances both flexibility and security.

\item The proposed framework incorporates a predictor-corrector mechanism that iteratively refines the recovery trajectory without external conditioning. This mechanism uses residual feedback from the current and predicted next steps to correct the reverse trajectory, which stabilizes latent representation extraction even for degraded stego image.

\item Extensive experiments demonstrate that the proposed framework outperforms state-of-the-art (SOTA) image steganography methods. Quantitative and qualitative results further show that the framework enables flexible control over stego image generation while preserving strong security against steganalysis. It also achieves superior recovery performance even in real transmission scenarios.
\end{enumerate}

The rest of this paper is organized as follows. Section~\ref{related} introduces some classical and diffusion-based generative image steganography. Section~\ref{methods} introduces the proposed method in detail. Section~\ref{Experiment} presents extensive experimental results and analysis. Section~\ref{conclusion} summarizes the paper.

\section{Related Works}\label{related}

\subsection{Modification-Based Image Steganography}
Traditional image steganography refers to the practice of concealing a secret image within a cover image in an unnoticed manner. Baluja~\cite{baluja2017hiding} pioneered the use of deep neural networks for image steganography and later extended this framework~\cite {baluja2019hiding} to conceal two secret images within a single cover image. Later studies incorporated GANs~\cite{goodfellow2014generative} into image steganography~\cite{Zhu2018ECCV}, where adversarial training is employed to embed a secret image into a cover image. The realistic appearance of the stego image ensures its security. SteganoGAN~\cite{zhang2019steganogan} introduces an adversarial network alongside the encoder and decoder networks for information embedding and extraction. This design enhances resistance to steganalysis. 

Recently, Invertible Neural Networks (INNs) have attracted attention in image steganography for establishing effective invertible mappings between secret and stego images~\cite{isn}. HiNet~\cite{hinet} models the image concealing and revealing as the forward and backward processes of an invertible network, which drastically increases both the hiding security and recovering accuracy. DeepMIH~\cite{guan2022deepmih} extends this paradigm to multi-image hiding by using the preceding stego image as guidance for embedding subsequent secret images. Subsequent works adopted different strategies to improve INN-based steganography. For instance, iSCMIS~\cite{iscmis} incorporates spatial and channel attention for secure region-guided embedding, while MIGIIHNet~\cite{migiihnet} employs mutual information estimation between the lost information and the stego image to achieve accurate recovery. However, these methods all rely on modifying the cover image to hide the secret image, which inevitably leaves detectable traces and renders the stego images vulnerable to well-designed steganalysis methods~\cite{xunet,kenet}. 

\subsection{Diffusion-Based Generative Image Steganography}
Diffusion models~\cite{song2020denoising, ho2020denoising} have recently emerged as a powerful generative framework for image generation \cite{hoogeboom2023simple, xia2023diffir, fei2023generative}. Among them, LDMs have attracted increasing attention in generative steganography~\cite{10877890, 10947094, 10902504,10659913} due to their superior image generation quality. Most existing methods rely on Denoising Diffusion Implicit Model (DDIM) inversion~\cite{song2020denoising} to construct the mapping from the secret image to the latent space, while the semantic content of the stego image is specified by text prompts.

\begin{figure*}[htbp]
  \centering
  \includegraphics[width=1\textwidth]{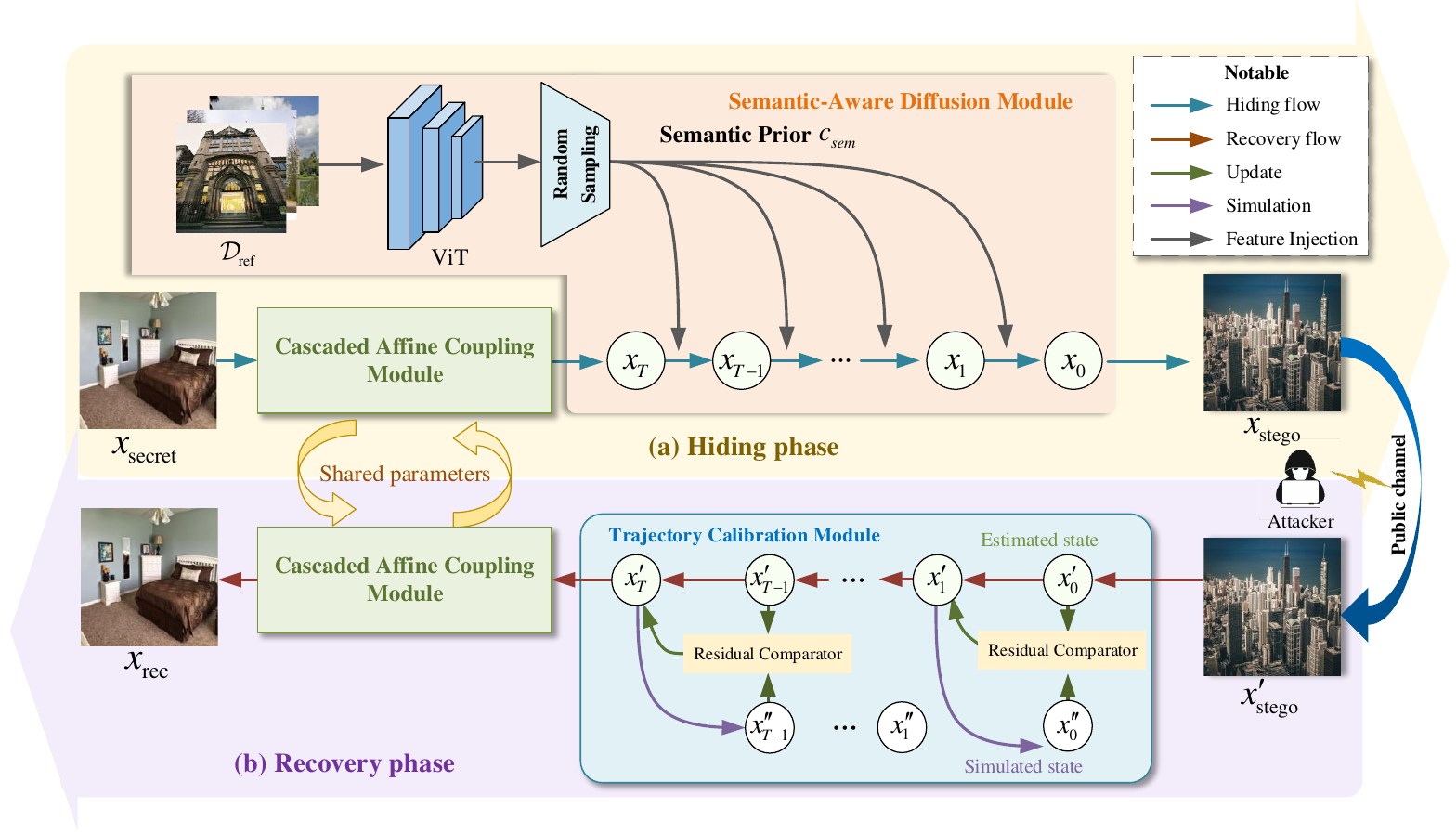}
  \caption{The proposed prompt-free generative steganography framework mainly consists of three modules: CACM, SADM, and TCM. These modules are responsible for secret image embedding/decoding, stego image generation, and latent recovery, respectively.}\label{framework}
\end{figure*}

CRoSS~\cite{yu2023cross} and subsequent works~\cite{xu2025diffusion} employ a text prompt to establish the mapping from the secret image to the latent space, while another text prompt is used to guide stego image generation. Although this mechanism ensures that the secret image can be recovered only with the correct prompts, text prompts have limited expressive ability for constructing the secret-to-latent mapping. Only part of the secret image can be concealed, which still compromises system security.
To improve the hiding capability in the latent space without relying on text prompts, subsequent studies introduce VQGAN compression~\cite{chen2025robust} and scrambling transformations~\cite {jiang2026image}. However, these methods still rely heavily on text prompts during stego image generation. Due to the limited ability of text prompts to describe complex visual details, the generated images may suffer from degraded visual quality\cite{zhang2023adding,xu2024prompt}. DiffStega~\cite{yang2024diffstega} introduces pre-determined passwords to generate reference images. These reference images provide auxiliary guidance for stego image generation and improve the visual details. However, the resulting stego images may contain content from both the reference and secret images. The significant content overlap between the secret and stego images, which compromises the confidentiality and the undetectability of the steganographic system.

In addition, stego images are often affected by lossy transmission in real-world scenarios, such as JPEG compression and Gaussian noise. These distortions may degrade the recovery accuracy. Existing work~\cite{yang2023provably} has focused mainly on robustness against known channel attacks. However, secret recovery in these methods still relies heavily on text prompts. Text prompts must be transmitted together with the stego image to guide secret recovery. During transmission, they are vulnerable to noise interference and tampering attacks~\cite{zhang2025provably}. Even minor perturbations to the transmitted text prompts can introduce errors into the recovery trajectory. Diffusion models exhibit inherent numerical instability and high sensitivity to perturbations~\cite{hu2024establishing}. Through iterative accumulation in the recovery process, these errors may cause the recovered trajectory to deviate significantly from the original trajectory~\cite{Wallace2023CVPR, Mokady2023CVPR}. Consequently, the reliance on text prompts poses a major challenge to robust recovery.

\section{Method}\label{methods}
As shown in Fig.~\ref{framework}, the proposed prompt-free generative image steganography framework consists of two stages. In the hiding phase, the secret image $x_{\text{secret}}$ is deterministically mapped by Cascaded Affine Coupling Module (CACM) to the latent representation $x_T$, which serves as the initial state of the diffusion process. In the Semantic-Aware Diffusion Module (SADM), a sampled semantic prior $c_{\text{sem}}$ extracted from $\mathcal{D}_{\text{ref}}$ by ViT is injected into the diffusion process to guide stego image generation. In the recovery phase, the received stego image $x'_{\text{stego}}$ is inverted through the Trajectory Calibration Module (TCM), which calibrates the recovery trajectory and obtains the latent representation $x'_T$. The recovered image $x_{\text{rec}}$ is reconstructed from the $x'_T$ through the inverse transformation of CACM.

\begin{figure}[htbp]
  \centering
  \includegraphics[width=1\linewidth]{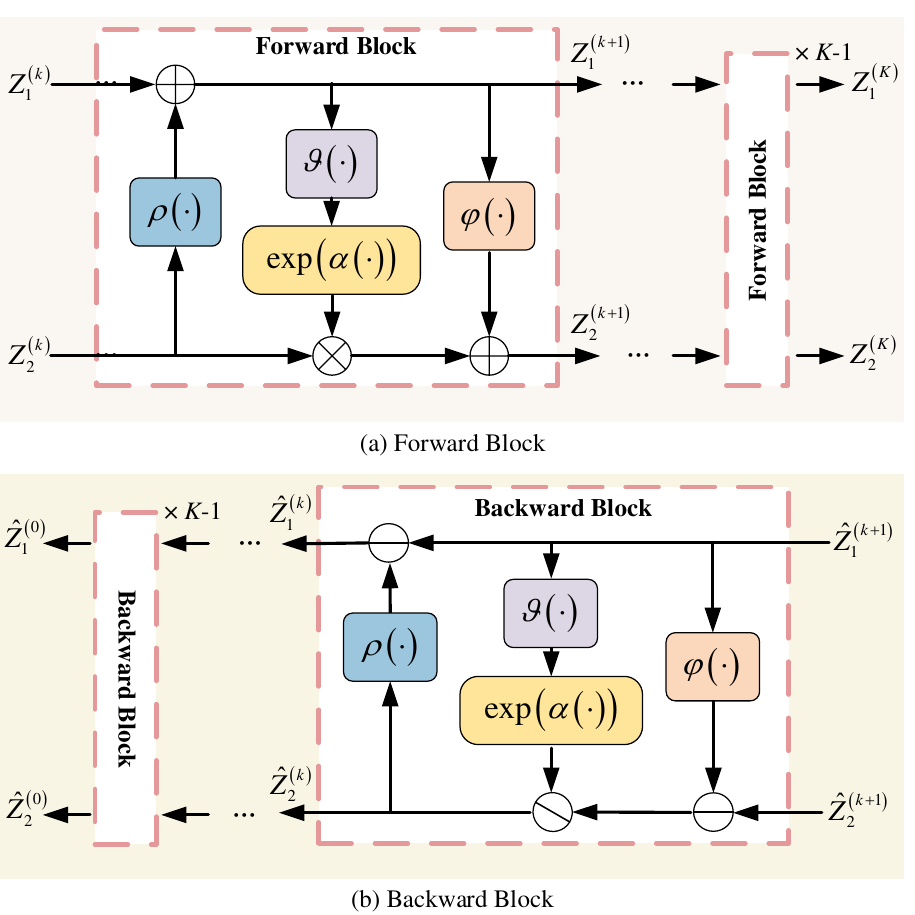}
  \caption{Architecture of the proposed CACM. The forward block maps the secret image to the latent representation, and the backward block reconstructs the image through the inverse transformation.}\label{CACM}
\end{figure}

\subsection{Cascaded Affine Coupling Module}
The CACM is designed to transform the secret image $x_{\text{secret}}$ into a latent representation $x_T$ that matches the input of the diffusion model. As shown in Fig.~\ref{CACM}, CACM comprises forward and backward blocks with shared parameters, which are used for secret embedding and recovery, respectively. The initial feature $Z^{(0)} = x_{\text{secret}}$ is first split into two parts along the channel dimension, denoted as $Z_1^{(0)}$ and $ Z_2^{(0)}$. In the $k$-th forward block, $Z_2^{(k-1)}$ first guides the update of $Z_1^{(k-1)}$. The updated feature $Z_1^{(k)}$ then generates the scale and shift terms to adaptively compress $Z_2^{(k-1)}$ through an affine coupling transformation. The forward block is formulated as follows:
\begin{equation}
    Z_1^{(k)} = Z_1^{(k-1)} \oplus \tau(Z_2^{(k-1)}),
\end{equation}
\begin{equation}
    Z_2^{(k)} = Z_2^{(k-1)} \odot \exp(\gamma(\vartheta(Z_1^{(k)}))) \oplus \eta(Z_1^{(k)}),
\end{equation}
where $\gamma(\cdot)$ denotes the sigmoid function. The operator $\odot$ denotes element-wise multiplication, and $\oplus$ denotes element-wise addition.
$\tau(\cdot)$, $\vartheta(\cdot)$, and $\eta(\cdot)$ denote three subnetworks that share the same architecture but do not share parameters. Specifically, $\tau(\cdot)$ is used to predict the additive update of $Z^{(k)}_1$. $\vartheta(\cdot)$ and $\eta(\cdot)$ are used to parameterize the scaling and shifting terms of the affine transformation on $Z^{(k)}_2$, respectively. Given the input $\phi_0$, the subnetwork is formulated as follows:
\begin{equation}
\begin{aligned}
&\mathcal{H}(\phi_0) = \mathrm{Concat}(\phi_0,\phi_1,\phi_2,\phi_3,\phi_4),\\
&\phi_i = \mathrm{LReLU}\!\left(\mathrm{Conv}_{3\times3}(\phi_{i-1})\right), \quad i=1,\ldots,4.
\end{aligned}
\end{equation}
where $\mathrm{Concat}(\cdot)$ denotes concatenation along the channel dimension. After the $K$-th forward block, the final latent representation $x_T$ is obtained by
\begin{equation}
x_T=\operatorname{Down}\big(\operatorname{Concat}(Z_1^{(K)},Z_2^{(K)})\big), \quad x_T \in \mathbb{R}^{4\times64\times64}
\end{equation}
where $\operatorname{Down}(\cdot)$ denotes the downsampling operation. $x_T$ is then used as the input to the subsequent diffusion model for stego image generation.

In the recovery process, the reconstructed latent representation $\hat{x}_T$ initializes the input feature $\hat{Z}^{(K)}$ of the $K$-th backward block. The subsequent propagation proceeds from the $(k+1)$-th block to the $k$-th block. The $k$-th backward block is formulated as follows:
\begin{equation}
    \hat{Z}_2^{(k-1)} = (\hat{Z}_2^{(k)} \ominus \eta(\hat{Z}_1^{(k)})) \oslash \exp(\gamma(\vartheta(\hat{Z}_1^{(k)}))),
\end{equation}
\begin{equation}
    \hat{Z}_1^{(k-1)} = \hat{Z}_1^{(k)} \ominus \tau(\hat{Z}_2^{(k-1)}),
\end{equation}
where $\oslash$ denotes element-wise division, and $\ominus$ represents element-wise subtraction. After the last backward block, the secret image is reconstructed as $\hat{Z}^{(0)} = x_\text{rec}$.

\subsection{Semantic-Aware Diffusion Module}
The SADM is designed to perform semantic-conditioned DDIM sampling from the latent representation $x_T$ to the stego image $x_{\text{stego}}$. Semantic priors are first extracted from reference images and injected into the prediction network $\epsilon_\theta$, which provides semantic control over the generating trajectory toward the target semantics.

Given a reference dataset $\mathcal{D}_{\text{ref}}$, a semantic prior set $\mathcal{C}_{\text{sem}}=\{\mathrm{ViT}(x)\mid x\in\mathcal{D}_{\text{ref}}\}$ is constructed by applying the ViT to the reference images. A semantic prior $c_{\text{sem}}$ is then randomly sampled from $\mathcal{C}_{\text{sem}}$ and injected into the prediction network $\epsilon_\theta$. $c_{\text{sem}}$ is fused with intermediate feature maps to provide semantic guidance and suppress potential artifacts.

The semantically guided sampling process can be formulated as the following ordinary differential equation (ODE):
\begin{equation}
    \frac{dx_t}{dt} = \left( f(t) + \frac{g^2(t)}{2\sigma_t^2} \right) x_t - \frac{\alpha_t g^2(t)}{2\sigma_t^2} \epsilon_\theta(x_t, t, c_{\text{sem}}),
\end{equation}
where $t\in [0,T]$, $f(t):=\frac{d \log \alpha_t}{dt}$, $g^2(t):= \frac{d \sigma^2_t}{dt}-2\frac{d \log \alpha_t}{dt} \sigma^2_t$. 

 At each step, DPM-Solver++~\cite{dpmsolver} approximates the exact solution at $t_i$ from the state $x_{t_{i-1}}$. This approximation is derived by applying a Taylor expansion around $\lambda_{t_{i-1}}$ as follows:
\begin{equation}
\begin{aligned}
x_{t_i} ={}& \frac{\sigma_{t_i}}{\sigma_{t_{i-1}}}x_{t_{i-1}} \\
&+ \sigma_{t_i} \sum_{n=0}^{w-1} \epsilon_\theta^{(n)}(x_{t_{i-1}}, t_{i-1}, c_{\text{sem}})
\int_{\lambda_{t_{i-1}}}^{\lambda_{t_i}} e^\lambda
\frac{(\lambda - \lambda_{t_{i-1}})^n}{n!} \, d\lambda \\
&+ \mathcal{O}(h_i^{w+1}).
\end{aligned}
\end{equation}

We use the first-order case ($w = 1$), and the corresponding update rule is written as follows:
\begin{equation}
\label{embde}
    x_{t_i} = \frac{\sigma_{t_i}}{\sigma_{t_{i-1}}}x_{t_{i-1}} - \alpha_{t_i}(e^{-h_i} - 1)\epsilon_\theta(x_{t_{i-1}}, t_{i-1}, c_{\text{sem}}),
\end{equation}
where $h_i = \lambda_{t_i} - \lambda_{t_{i-1}}$ represents the step size in the log-SNR space.

The secret latent representation $x_T$ is used as the initial state of the generation process, and the stego image $x_{\text{stego}}$ is progressively generated from it. Algorithm~\ref{alg:stego_gen} summarizes the stego image generation procedure of the proposed method.

\begin{algorithm}[t]
\caption{Stego Image Generation with Semantic Guidance}
\label{alg:stego_gen}
\begin{algorithmic}
\STATE \textbf{Input:} Secret image $x_{\text{secret}}$, reference dataset $\mathcal{D}_{\text{ref}}$, number of affine coupling blocks $K$, time steps $\{t_i\}_{i=1}^{N}$.\\
\STATE \textbf{Output:} Stego image $x_{\text{stego}}$.
\STATE
\STATE \textbf{Stage 1: Invertible latent mapping and encoding}
\STATE Initialize $Z^{(0)} \gets x_{\text{secret}}$, and split $Z^{(0)}$ into $[Z_1^{(0)}, Z_2^{(0)}]$.
\FOR{$k = 1, \ldots, K$}
    \STATE $Z_1^{(k)} \gets Z_1^{(k-1)} \oplus \tau\left(Z_2^{(k-1)}\right)$.
    \STATE $Z_2^{(k)} \gets Z_2^{(k-1)} \odot \exp\left(\gamma\left(\vartheta\left(Z_1^{(k)}\right)\right)\right) \oplus \eta\left(Z_1^{(k)}\right)$.
\ENDFOR
\STATE Assemble secret latent $\hat{x}_T \gets [Z_1^{(K)}, Z_2^{(K)}]$.
\STATE
\STATE \textbf{Stage 2: Semantic prior sampling}
\STATE $\mathcal{C}_{\text{sem}} \gets \operatorname{ViT}\left(\mathcal{D}_{\text{ref}}\right)$.
\STATE $c_{\text{sem}} \gets \operatorname{RandomSample}\left(\mathcal{C}_{\text{sem}}\right)$.
\STATE
\STATE \textbf{Stage 3: Semantic-conditioned diffusion generation}
\STATE $x_{t_N} \gets \hat{x}_T$.
\FOR{$i = N, \ldots, 1$}
    \STATE $\hat{\epsilon}_i \gets \epsilon_{\theta}\left(x_{t_i}, t_i, c_{\text{sem}}\right)$.
    \STATE $h_i \gets \lambda_{t_{i-1}} - \lambda_{t_i}$.
    \STATE $x_{t_{i-1}} \gets \operatorname{DPM\text{-}Solver++}\left(x_{t_i}, \hat{\epsilon}_i, h_i\right)$.
\ENDFOR
\STATE $x_{\text{stego}} \gets \operatorname{Decode}\left(x_{t_0}\right)$.
\STATE \textbf{return} $x_{\text{stego}}$.
\end{algorithmic}
\end{algorithm}

\subsection{Trajectory Calibration Module}
In the recovery stage, the stego image $x_{\text{stego}}$ is inverted back to the latent representation without semantic conditioning. From Eq.~\eqref{embde}, the inversion at time $t_i$ can be written as follows:
\begin{equation}
    x'_{t_i} = \frac{\sigma_{t_i}}{\sigma_{t_{i-1}}} x'_{t_{i-1}} - \alpha_{t_i} (e^{-h_i} - 1) \epsilon_\theta(x'_{t_{i-1}}, t_{i-1}).
\end{equation}

Since $x'_{t_{i-1}}$ is unknown during the current step, the prediction term $\epsilon_\theta(x'_{t_{i-1}}, t_{i-1})$ cannot be directly obtained. In na\"{\i}ve DDIM inversion, $x'_{t_i}$ is used to approximate the unknown term $x'_{t_{i-1}}$, which inevitably introduces numerical errors. To address this issue, we introduce the TCM that predicts the next state and then calibrates the recovery trajectory through single-step discrepancy feedback. The TCM first computes an estimate state $x'_{t_{i-1}}$ from the current state $x'_{t_i}$ using na\"{\i}ve DDIM inversion:
\begin{equation}
    x'_{t_{i-1}} = \frac{\sigma_{t_{i-1}}}{\sigma_{t_i}} \left( x'_{t_i} + \alpha_{t_i} (e^{-h_i} - 1) \epsilon_\theta(x'_{t_i}, t_i) \right).
\end{equation}
The estimated state $x'_{t_{i-1}}$ is used to construct a simulated forward state $x''_{t_i}$:
\begin{equation}
    x''_{t_i} \leftarrow \frac{\sigma_{t_i}}{\sigma_{t_{i-1}}} x'_{t_{i-1}} - \alpha_{t_i} (e^{-h_i} - 1) \epsilon_\theta(x'_{t_{i-1}}, t_{i-1}).
\end{equation}
The mismatch between the simulated state $x''_{t_i}$ and the observed current state $x'_{t_i}$ is measured by:
\begin{equation}
    \mathcal{R}_i=\|x''_{t_i}-x'_{t_i}\|_2^2.
\end{equation}
The estimated state $x'_{t_{i-1}}$ is then refined by using the gradient of $\mathcal{R}_i$:
\begin{equation}
    x'_{t_{i-1}} \leftarrow x'_{t_{i-1}}-\vartheta \nabla_{x'_{t_{i-1}}}\mathcal{R}_i,
\end{equation}
where $\vartheta$ denotes the update step size for the gradient trajectory calibration. This operation performs gradient correction based on the mismatch between the forward-projected state and the current state. In this way, TCM not only compensates for the approximation error of DDIM inversion but also corrects reverse trajectory deviations caused by distortions in the transmission channel. The recovered latent representation $\hat{x}_T$ is then fed into CACM for accurate reconstruction of the secret image $x_{\text{rec}}$. Algorithm~\ref{alg:SecretRecovery} provides a detailed outline of the comprehensive steps involved in the proposed method for recovering the secret image.

\begin{algorithm}[t]
\caption{Secret Image Recovery without Condition}
\label{alg:SecretRecovery}
\begin{algorithmic}

\STATE \textbf{Input:} Stego image $x_{\text{stego}}$, inversion steps $N$, time steps $\{t_i\}_{i=1}^{N}$, number of affine coupling blocks $K$.
\STATE \textbf{Output:} Recovered secret image $x_{\text{rec}}$.

\STATE
\STATE \textbf{Stage 1: Unconditional inversion with TCM}
\STATE $x'_{t_0} \gets x_{\text{stego}}$

\FOR{$i=1,\ldots,N$}
    \STATE $x'_{t_{i-1}} \gets
    \frac{\sigma_{t_{i-1}}}{\sigma_{t_i}}
    \left(
    x'_{t_i}
    +
    \alpha_{t_i}(e^{-h_i}-1)
    \epsilon_{\theta}(x'_{t_i},t_i)
    \right)$

    \REPEAT
        \STATE $x''_{t_i} \gets
        \frac{\sigma_{t_i}}{\sigma_{t_{i-1}}}x'_{t_{i-1}}
        -
        \alpha_{t_i}(e^{-h_i}-1)
        \epsilon_{\theta}(x'_{t_{i-1}},t_{i-1})$

        \STATE $x'_{t_{i-1}} \gets x'_{t_{i-1}}
        -
        \vartheta \nabla_{x'_{t_{i-1}}}\mathcal{R}_i$

        \STATE $\operatorname{UPDATE}(x'_{t_{i-1}},x''_{t_i},x'_{t_i})$
    \UNTIL{converged}
\ENDFOR

\STATE $\hat{x}_T \gets x'_{t_N}$

\STATE
\STATE \textbf{Stage 2: Inverse latent mapping and decoding}
\STATE Split $\hat{x}_T$ into $[\hat{Z}_1^{(K)},\hat{Z}_2^{(K)}]$

\FOR{$k=K,\ldots,1$}
    \STATE $\hat{Z}_2^{(k-1)} \gets
    (\hat{Z}_2^{(k)} \ominus \eta(\hat{Z}_1^{(k)}))
    \oslash
    \exp(\gamma(\vartheta(\hat{Z}_1^{(k)})))$

    \STATE $\hat{Z}_1^{(k-1)} \gets
    \hat{Z}_1^{(k)} \ominus \tau(\hat{Z}_2^{(k-1)})$
\ENDFOR

\STATE $x_{\text{rec}} \gets
\operatorname{Concat}(\hat{Z}_1^{(0)},\hat{Z}_2^{(0)})$

\STATE \textbf{return} $x_{\text{rec}}$

\end{algorithmic}
\end{algorithm}

\subsection{Loss Functions}
We combine the loss terms introduced to form the total loss function $\mathcal{L}_{\text{total}}$. This objective is composed of a recovery loss $\mathcal{L}_{\text{rec}}$ and a trajectory consistency loss $\mathcal{L}_{\text{tc}}$, which are defined as follows:
\begin{equation}
    \mathcal{L}_{\text{total}} = \lambda_{\text{rec}}\mathcal{L}_{\text{rec}} + \lambda_{\text{tc}}\mathcal{L}_{\text{tc}},
\end{equation}
where $\lambda_{\text{rec}}$ and $\lambda_{\text{tc}}$ are weighted coefficients for training.

\paragraph{Recovery Loss} 
To ensure accurate reconstruction of the secret image during recovery, we introduce a recovery loss $\mathcal{L}_{\text{rec}}$, which minimizes the discrepancy between the recovery image $x_{\text{rec}}$ and the secret image $x_{\text{secret}}$. The loss is defined as:
\begin{equation}
    \mathcal{L}_{\text{rec}} = \| x_{\text{rec}} - x_{\text{secret}} \|_1,
\end{equation}
where $\| \cdot \|_1$ denotes the $L_1$ norm regularization.

\paragraph{Trajectory Consistency Loss} 
To ensure accurate recovery, we introduce a trajectory consistency loss that measures the discrepancy between states. The loss computes the squared $L_2$ distance between the original state and the simulated state $\hat{x}_T$ obtained through reverse inversion:
\begin{equation}
    \mathcal{L}_{\text{tc}} = \| x_T - \hat{x}_T \|_2^2.
\end{equation}

\section{Experiment}\label{Experiment}
In this section, we first introduce the experimental settings.
Then, we provide a quantitative and qualitative evaluation on the visual appearance of the generated stego images. The security performance against some steganalyzers is reported also. Additionally, we assess its robustness against different degradations. Finally, ablation experiments are conducted to illustrate the necessity of each component in our framework.

\subsection{Implementation Details}
\paragraph{Experimental Settings} 
In our experiments, Stable Diffusion v2~\cite{rombach2022high} provided by HuggingFace is selected as the baseline. The guidance scale of the diffusion model is set to 3.0. During inference, DPMSolver++~\cite{dpmsolver} is used for sampling with 50 steps. 

The number of CACM layers is set to $K=4$. The hyperparameters of the loss function $\lambda_{\text{rec}}$ and $\lambda_{\text{tc}}$ are set to 2.0 and 1.0, respectively. The model is trained for 20 epochs with the Adam optimizer and a learning rate of $1.0 \times 10^{-4}$. The batch size is set to 1. All experiments are conducted on an NVIDIA L20 GPU with the PyTorch framework.
\paragraph{Baseline Methods}
We focus on high-capacity image steganography and select seven classical baseline methods for comparison. These methods include some traditional modification-based image steganography (TS) methods, such as HiNet~\cite{hinet}, ISN~\cite{isn}, iSCMIS~\cite{iscmis}, and MIGIIHNet~\cite{migiihnet}. State-of-the-art generative steganography (GS) methods, including CRoSS~\cite{yu2023cross}, DiffStega~\cite{yang2024diffstega}, and VQGAN-DiffStega~\cite{chen2025robust}. All these models
utilized in this study were pre-trained.

\paragraph{Data Preparation} 
Following the data setting in~\cite{chen2025robust}, we use the FFHQ dataset~\cite{Karras2019CVPR} as the secret image set, with 5,000 images for training and 1,000 for testing. For GS methods that require text prompts, we use ChatGPT to extract text prompts from the Bedroom category of the LSUN dataset~\cite{lsun}. For TS methods, the cover images are generated by the same diffusion model. All images are center-cropped to $512 \times 512$ pixels and the corresponding latent space has a dimension of $64 \times 64 \times 4$.  The embedding capacity for all baselines is set at 24 bpp (bits per pixel).

\paragraph{Evaluation metrics} 
In the experiments, we adopt various metrics to evaluate the performance of diverse steganographic methods. Specifically, we measure extraction accuracy, robustness, security, and visual quality. 

\textbf{Visual quality:} The Fréchet Inception Distance (FID) \cite{heusel2017gans} serves as a benchmark for assessing image quality, as it quantifies the Fréchet distance between the feature distributions of stego images and real images. The FID is defined as follows:
\begin{equation}
\mathrm{FID}_{(x,y)}=\|\mu_x-\mu_y\|^2+\mathrm{Tr}\left(\Sigma_x+\Sigma_y-2\left(\Sigma_x\Sigma_y\right)^{1/2}\right),
\end{equation}
where $\mu_x$ and $\mu_y$ denote the mean feature vectors of the real and stego images, respectively. $\Sigma_x$ and $\Sigma_y$ denote their corresponding covariance matrices. 

For security, the stego image should be visually authentic and with no content leakage. According to this, we calculate two FIDs, i.e. $\mathrm{FID}_{(\text{ste},\text{real})}$ and $\mathrm{FID}_{(\text{ste},\text{sec})}$. The former measures the distribution distance between stego images and real images (bedroom dataset~\cite{lsun}) to assess authenticity, for which a lower score indicates higher resemblance to real images. For the latter, we use the secret FFHQ dataset as the real image set. A higher score implies less information leakage.

Natural Image Quality Evaluator (NIQE)\cite{zhang2015feature} is a no-reference image quality assessment metric that evaluates the perceptual naturalness of stego images, with lower scores indicating greater statistical similarity to natural images.

\textbf{Extraction accuracy:} We use PSNR and SSIM to measure extraction accuracy by comparing the recovered image with the original secret image, where higher values indicate better reconstruction. PSNR is defined based on the mean squared error (MSE) as follows:
\begin{equation}
\mathrm{MSE} =\frac{1}{W\times H} \sum_{i=1}^{W} \sum_{j=1}^{H}\left[X_{i, j}-Y_{i, j}\right]^{2},
\end{equation}
\begin{equation}
    \text { PSNR }  =10 \cdot \log _{10} \frac{\mathrm{MAX}^{2}}{\mathrm{MSE}}.
\end{equation}

SSIM is defined as follows:
\begin{equation}
\operatorname{SSIM}=\frac{\left(2 \mu_{x} \mu_{y}+c_{1}\right)\left(2 \sigma_{xy}+c_{2}\right)}{\left(\mu_{x}^{2}+\mu_{y}^{2}+c_{1}\right)\left(\sigma_{x}^{2}+\sigma_{y}^{2}+c_{2}\right)}.
\end{equation}

\textbf{Security:} The security refers the detection resistance of stego image against the steganalysis tool. It is commonly evaluated by the detection error $P_E$ of the steganalyzer, which is defined as follows:
\begin{equation}
P_E = \frac{ P_{\mathrm{FA}} + P_{\mathrm{MD}}}{2} ,
\end{equation}
where \( P_{\mathrm{FA}} \) is the false alarm rate and \( P_{\mathrm{MD}} \) is the missed detection rate. The value of $P_E$ ranges from 0 to 1. $P_E=0.5$ corresponds to random guessing (equal likelihood of misclassifying cover or stego images), indicating optimal resistance to steganalysis. To measure $P_E$, we adopt three steganalysis tools: StegExpose~\cite{stegexpose}, XuNet~\cite{xunet} and KeNet~\cite{kenet}. StegExpose is a widely used steganalysis tool that integrates multiple statistical methods, including RS analysis and the Chi-square attack. While XuNet and KeNet are representative deep learning-based steganalyzers.

\textbf{Robustness:} To evaluate robustness against unknown real-world distortions, we test extraction accuracy under several attacks, including JPEG compression, Gaussian noise, Salt-and-pepper noise, Gaussian filtering, Median filtering, rotation, and resizing.

\subsection{Visual Quality Comparison}
Since traditional methods(TS) preserve visual quality by modifying a cover image, visual comparison with GS methods on visual quality is unfair. Therefore, we focus on the visual quality comparison among GS methods. 

\begin{figure}[htbp]
  \centering
  \includegraphics[width=1\linewidth]{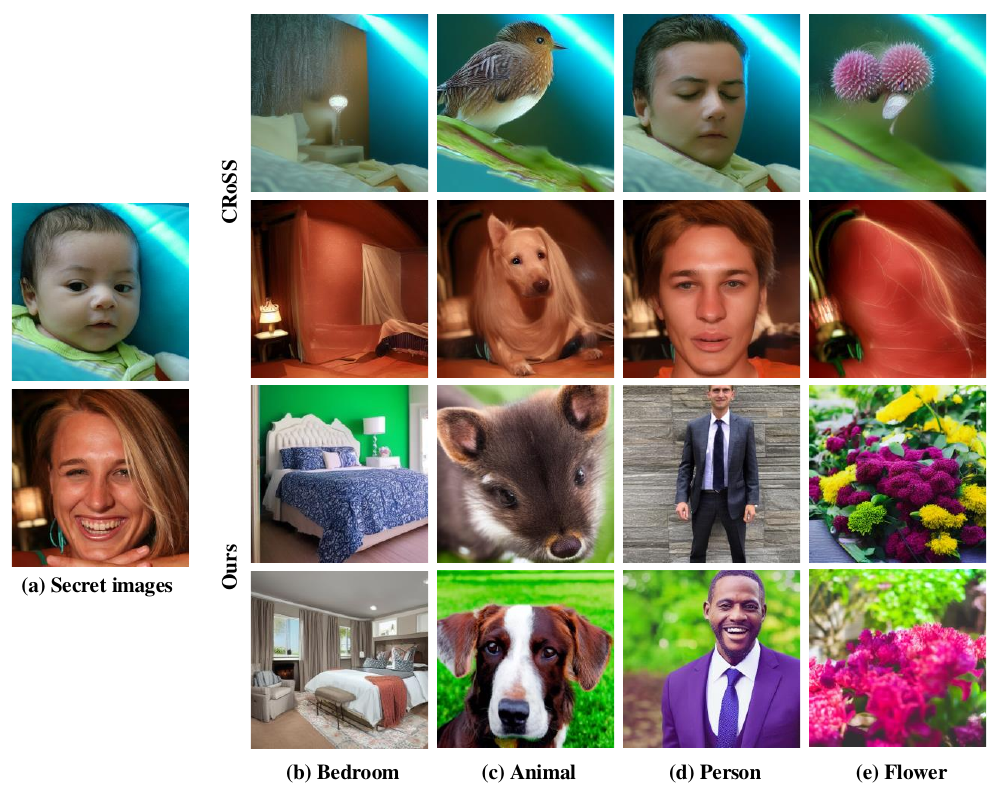}
  \caption{Stego images generated by CRoSS and our proposed method with the same semantic conditions. }\label{fig:stego}
\end{figure}

Figure~\ref{fig:stego} shows stego images generated by CRoSS and the proposed method under four semantic conditions: \textquotesingle Flower\textquotesingle, \textquotesingle Bedroom\textquotesingle, \textquotesingle Animal\textquotesingle, and \textquotesingle Human\textquotesingle. Each column shares the same semantic condition. The first two rows show stego images from CRoSS, guided by label names as text prompts. Since text prompts lack the ability to capture fine-grained details, the stego images suffer from reduced visual authenticity and content leakage. The third and fourth rows present our method guided by semantic priors from reference images. It is obvious that the proposed method produce a diverse set of images from the same secret image. The obtained stego images exhibit clearer textures, and a better semantic alignment compared to the obtained from the other approaches.

For quantitative comparison, we further adopt NIQE and $\mathrm{FID}_{(\text{ste},\text{real})}$ to evaluate the visual quality of our method. The results are given in the first two columns of Table~\ref{tab:visual_comparison}. Compared with CRoSS~\cite{yu2023cross} and DiffStega~\cite{yang2024diffstega}, VQGAN-DiffStega~\cite{chen2025robust} shows relatively competitive performance ob both two metrics. The proposed method achieves the lowest NIQE and $\mathrm{FID}_{(\text{ste},\text{real})}$ scores, which further confirms its superior visual quality.

\begin{table}
\centering
\caption{The visual comparison of stego images generated by different generative-based schemes. The best results are in bold and the second-best results are underlined.}
\label{tab:visual_comparison}
\begin{tabular}{lccc} 
\hline
Methods         & NIQE $\downarrow$ & $\mathrm{FID}_{(\text{ste},\text{real})}$ $\downarrow$ & $\mathrm{FID}_{(\text{ste},\text{sec})}$ $\uparrow$     \\ 
\hline
CRoSS~\cite{yu2023cross}           & 4.42              & 49.052           & 71.608   \\
DiffStega~\cite{yang2024diffstega}       & 4.966             & 48.698           & 74.055   \\
VQGAN-DiffStega~\cite{chen2025robust} & \underline{3.25}      & \underline{17.709}   & \underline{292.933}  \\
Ours            & \textbf{3.15}     & \textbf{9.254}   & \textbf{342.506}  \\
\hline
\end{tabular}
\end{table}

\subsection{Security Analysis}
We compare the proposed framework with several state‑of‑the‑art (SOTA) methods to evaluate its security in terms of imperceptibility and undetectability. Imperceptibility is determined by how much secret-related content remains in the stego image. Following \cite{chen2025robust}, we quantify this effect by measuring the distribution distance between the secret and stego image sets. As shown in the third column of Table~\ref{tab:visual_comparison}, CRoSS~\cite{yu2023cross} and DiffStega~\cite{yang2024diffstega} obtain lower $\mathrm{FID}_{(\text{ste},\text{sec})}$ scores. This indicates a higher resemblance between the secret and stego image sets, which implies a severe risk of semantic leakage. By contrast, the proposed framework overcomes this issue by aligning the generation with the target semantics. As expected, the proposed method achieves the highest $\mathrm{FID}_{(\text{ste},\text{sec})}$ score, which suggests the lowest risk of content leakage.

\begin{figure}[htbp]
  \centering
  \includegraphics[width=1\linewidth]{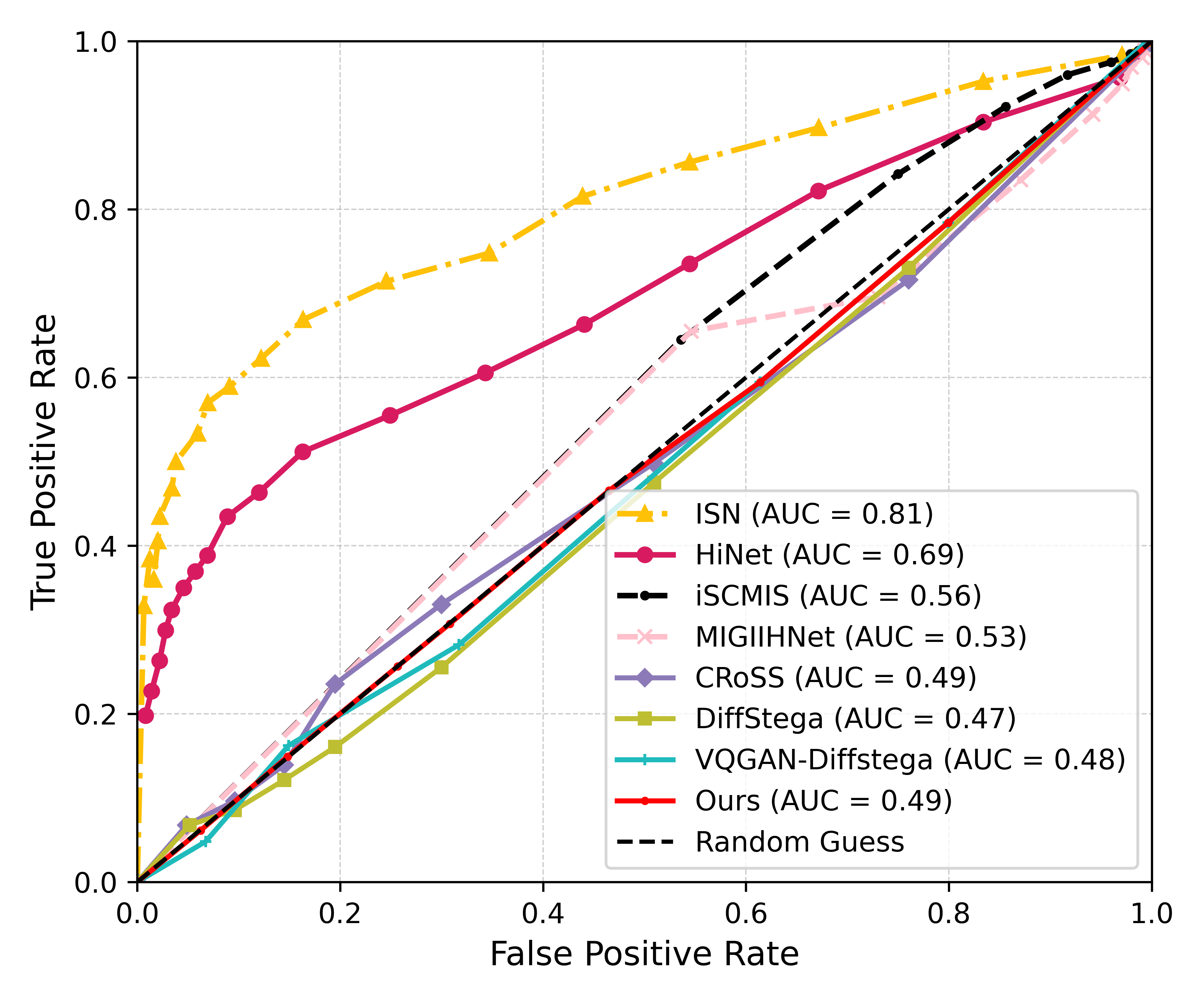}
  \caption{ ROC curves of StegExpose for different schemes.}\label{fig:roc}
\end{figure}

To evaluate the undetectability of the proposed method, 2,000 cover/stego image pairs generated by state-of-the-art (SOTA) methods were utilized for steganalysis. For fair comparison, the cover images are generated by the same diffusion model without embedding any secret. These cover images serve both as carriers for TS methods and as a reference for GS‑generated stego images.

\begin{table}[htbp] 
\centering
\caption{Quantitative comparison of undetectability among different methods. The best results are in bold and the second-best results are underlined.}
\label{tab:steganalysis}
\resizebox{1\linewidth}{!}{
\begin{tabular}{lccc}
\toprule
Methods & Type  &  XuNet \cite{xunet} & KeNet \cite{kenet} \\ 
        \midrule
        HiNet \cite{hinet} & \multirow{4}{*}{TS}  & 0.5591 & 0.5833 \\
        ISN \cite{isn} &    & 0.5512 & 0.5827 \\
        iSCMIS \cite{iscmis} &    & 0.4755 & 0.5483 \\
        MIGIIHNet \cite{migiihnet} &    & 0.4910 & 0.5163 \\ 
        \midrule
        CRoSS \cite{yu2023cross} & \multirow{4}{*}{GS}   & 0.5132 & 0.5211 \\
        DiffStega \cite{yang2024diffstega} &   & 0.5078 & - \\
        VQGAN-DiffStega \cite{chen2025robust} &   & 0.5198 & - \\
        \textbf{Ours} &   & \textbf{0.5005} & \textbf{0.5049} \\ 
\bottomrule
\end{tabular}
}
\end{table}
Fig.~\ref{fig:roc} shows the receiver operating characteristic (ROC) curves for StegExpose~\cite{stegexpose} for different methods. The proposed method achieves an area under the ROC curve (AUC) of 0.49, which is close to the level of a random classifier. This indicates the StegExpose cannot reliably distinguish the generated stego images from cover images,  which confirms the superior undetectability of our approach. Furthermore, the detection error rates obtained by XuNet and KeNet are also reported in Table~\ref{tab:steganalysis}. As shown in the table, stego images generated by TS methods, such as HiNet~\cite{hinet} and ISN~\cite{isn}, are more easily detectable, as modifying cover images inevitably introduces statistical artifacts. Unlike TS methods, GS methods synthesize stego images directly without any cover modification, which minimizes detectable artifacts and offers stronger resistance against steganalysis. Notably, the detection error rate of the proposed method remains consistently closer to 0.5 than the compared methods, indicating its superior undetectability.

\subsection{Robustness}
\begin{figure*}[htbp]
  \centering
  \includegraphics[width=0.9\textwidth]{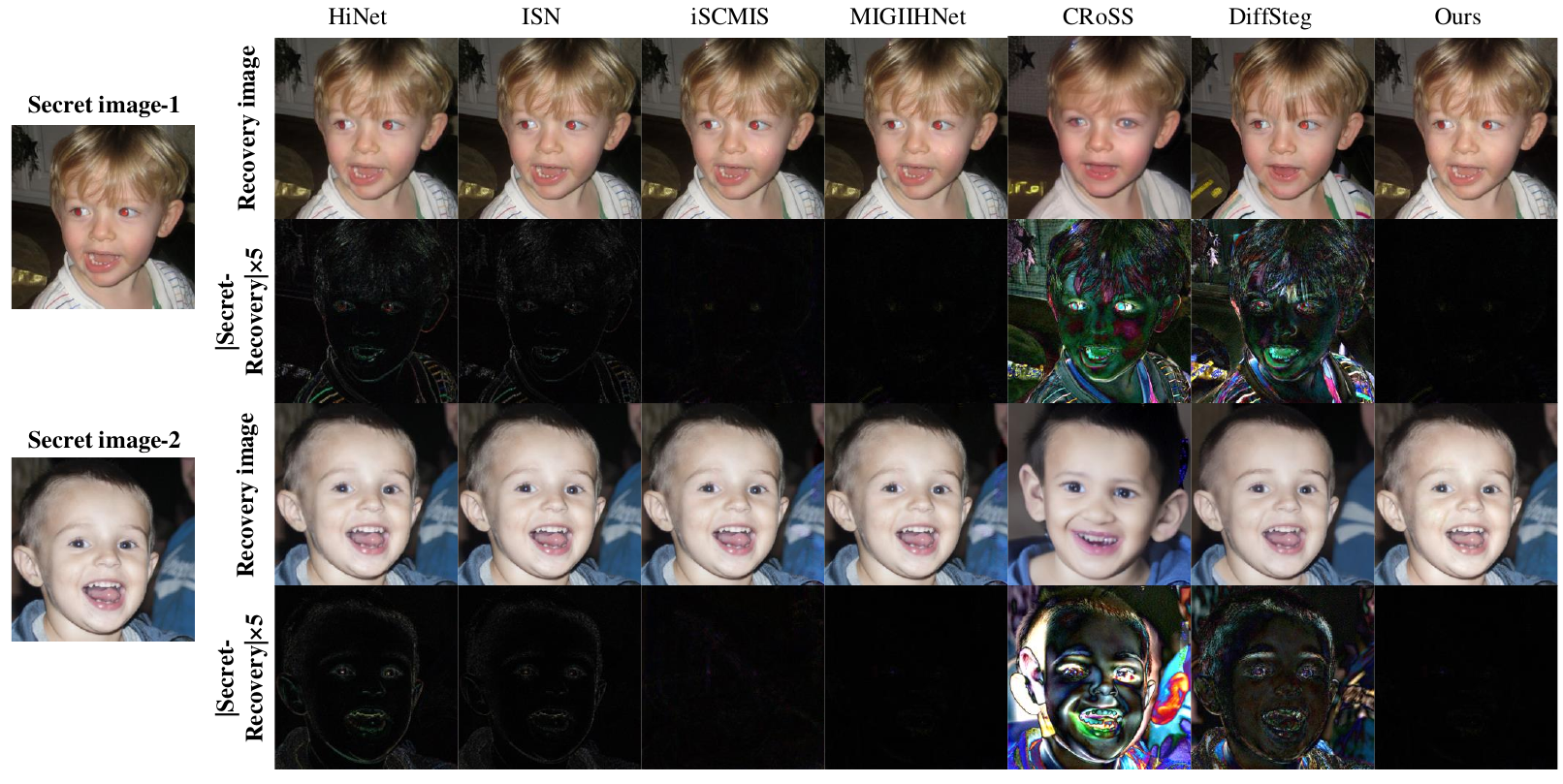}
  \caption{Demonstrations of the recovery images obtained by different schemes.}\label{recoverey}
\end{figure*}

To evaluate robustness, we simulate stego image degradation using Gaussian noise and JPEG compression at different levels. Table~\ref{tab:robustness} reports the PSNR and SSIM scores of the recovered images. The first column of the table shows that TS methods outperform standard GS methods in recovering secret images from clean stego images. Our method achieves competitive performance, a conclusion further supported by the visual comparison of recovered images in Fig.~\ref{recoverey}. 

\begin{table*}[htbp]
\centering
\caption{PSNR/SSIM of recovered images from degraded stego images. The best results are in bold and the second-best results are underlined.}
\label{tab:robustness}
\resizebox{1\linewidth}{!}{
\begin{tabular}{lcccccccccc}
\toprule
\multirow{2}{*}{Methods} & \multirow{2}{*}{Type} & \multirow{2}{*}{Clean} & \multicolumn{4}{c}{Gaussian Noise} & \multicolumn{4}{c}{JPEG Compression} \\
\cmidrule(lr){4-7} \cmidrule(lr){8-11}
& & & $\sigma=0.035$ & $\sigma=0.025$ & $\sigma=0.015$ & $\sigma=0.005$ & QF=30 & QF=50 & QF=70 & QF=90 \\ \hline
HiNet \cite{hinet} & \multirow{4}{*}{TS} & 37.05/0.93 & 10.64/0.05 & 12.72/0.09 & 16.46/0.18 & 19.65/0.28 & 11.16/0.32 & 11.23/0.32 & 11.25/0.33 & 11.30/0.33 \\
ISN \cite{isn} & & 36.77/0.96 & 13.42/0.12 & 16.04/0.19 & 17.85/0.24 & 20.19/0.33 & 8.30/0.38 & 8.61/0.40 & 8.75/0.40 & 8.82/0.42 \\
iSCMIS \cite{iscmis} & & 44.71/0.98 & 13.45/0.13 & 19.88/0.25 & 21.30/0.34 & 26.83/0.57 & 11.34/0.44 & 11.30/0.38 & 11.27/0.34 & 11.34/0.27 \\
MIGIIHNet \cite{migiihnet} & & \textbf{46.76/0.99} & 15.44/0.20 & 18.47/0.25 & 20.63/0.27 & 28.27/0.33 & 11.51/0.38 & 11.47/0.37 & 11.48/0.35 & 11.47/0.32 \\ \hline
CRoSS \cite{yu2023cross} & \multirow{4}{*}{GS} & 22.75/0.74 & 20.98/0.51 & 21.55/0.58 & 22.28/0.65 & 22.77/0.68 & 20.36/0.58 & 21.01/0.60 & 22.01/0.67 & 22.54/0.70 \\
DiffStega \cite{yang2024diffstega} & & 23.13/0.74 & 21.08/0.52 & 21.66/0.59 & 22.34/0.65 & 22.67/0.68 & 20.57/0.60 & 21.23/0.63 & 22.13/0.69 & 22.64/0.71 \\
VQGAN-DS \cite{chen2025robust} & & 23.65/0.72 & \underline{21.20/0.63} & \underline{21.74/0.64} & \underline{22.47/0.66} & \underline{23.09/0.69} & \underline{20.86/0.62} & \underline{21.39/0.64} & \underline{22.24/0.67} & \underline{22.88/0.69} \\
\textbf{Ours} & & \underline{46.50/0.99} & \textbf{33.29/0.97} & \textbf{33.33/0.97} & \textbf{33.23/0.97} & \textbf{33.14/0.97} & \textbf{37.34/0.99} & \textbf{38.24/0.99} & \textbf{39.35/0.99} & \textbf{40.75/0.99} \\ \bottomrule
\end{tabular}
}
\end{table*}

\begin{figure*}
  \centering
  \includegraphics[width=1\textwidth]{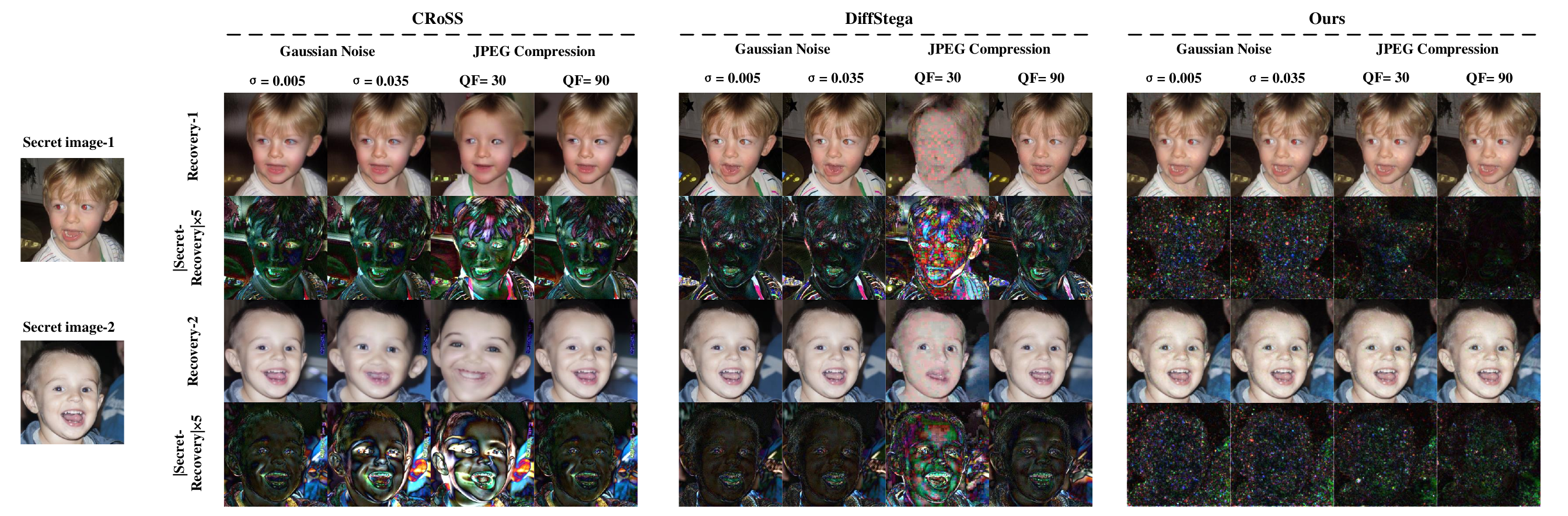}
  \caption{Visual comparison of the recovered image under different levels of degradation.}\label{noise}
\end{figure*}

However, for degraded stego images, TS performance degrades significantly while GS methods including CRoSS, DiffStega and VQGAN‑DS demonstrate stronger resilience and yield relatively consistent recovery accuracy. The proposed method achieves the best and most stable recovery scores across all degradation levels. Fig.~\ref{noise} provides a visual comparison, including the original secret images and their corresponding recovered versions from stego images degraded at different levels. Our scheme demonstrates remarkable fidelity in recovering the secret content, whereas images recovered by other baseline schemes exhibit noticeable color and texture distortions.

\subsection{Ablation Studies}
\begin{table*}[htbp]
\centering
\caption{Quantitative results of the ablation study on different configurations.}\label{tab:ablation}
\begin{tabular}{ccccccc} 
\hline
\multirow{2}{*}{Setting} & \multirow{2}{*}{NIQE $\downarrow$} & \multirow{2}{*}{$\mathrm{FID}_{(\text{ste},\text{real})}$ $\downarrow$} & \multicolumn{2}{c}{Secret/Recovery image pairs} & \multirow{2}{*}{Detection error(\%)} & \multirow{2}{*}{$\mathrm{FID}_{(\text{ste},\text{sec})}$ $\uparrow$}  \\ 
\cline{4-5}
                         &                                    &                                                                         & PSNR $\uparrow$ & SSIM $\uparrow$               &                                      &                                                                       \\ 
\hline
Baseline                 & 4.42                  & 49.052               & 22.75 & 0.74                                    & 0.52                                 & 71.608                \\
Baseline + CACM          & 4.38                  & 48.292               & 32.89 & 0.96                                    & 0.52                                 & 75.698                \\
Baseline + TCM           & 4.40                  & 48.885               & 34.65 & 0.97                                    & 0.47                                 & 72.328                \\
Baseline + SADM + TCM    & 3.58                  & 18.712               & 35.12 & 0.97                                    & 0.48                                 & 278.57                \\
Ours                     & 3.07                  & 8.670                & 46.50  & 0.99                                    & 0.51                                 & 342.51               \\
\hline
\end{tabular}
\end{table*}
To assess the contribution of each component in our framework, we conduct ablation experiments by adopting CRoSS~\cite{yu2023cross} as the baseline. The CRoSS result is taken from the reported result in~\cite{chen2025robust}.
The proposed modules are then progressively integrated into this baseline, and the resulting variants are evaluated under the same experimental setting. 
\paragraph{Effectiveness of the CACM} CACM is employed to embed secret image into the latent representation. Rows 1 and 2 of the Table~\ref{tab:ablation} report the results of baseline and CACM‑enabled configuration, respectively.  Compared to the baseline, the integration of CACM substantially boosts reconstruction fidelity, which increases the PSNR from 22.75 dB to 32.89 dB and the SSIM from 0.74 to 0.96. These gains indicate that CACM establishes a stable and deterministic mapping between the secret image and the latent representation,  which is a critical property for reliable information embedding.

\paragraph{Effectiveness of the SADM} The proposed SADM is used to guide stego image generation with style priors. The  efficiency of SADM is demonstrated by comparing the configurations shown in Rows 3 and 4 of Table~\ref{tab:ablation}. With SADM, the visual quality of the stego images is significantly improved,  whose $\mathrm{FID}_{(\text{ste},\text{real})}$ drops from 41.885 to 18.712. This improvement is attributed to the semantic priors introduced by SADM, which enhance the undetectability of the generated stego images.

\begin{table*}[htbp]
\centering
\caption{Ablation study on the inversion strategy under different distortion levels}
\label{tab:ablation_inver}
\resizebox{1\textwidth}{!}{
\begin{tabular}{cccccccccc} 
\hline
\multirow{2}{*}{Method} & \multirow{2}{*}{Clean} & \multicolumn{4}{c}{Gaussian Noise}                & \multicolumn{4}{c}{JPEG compression}               \\ 
\cline{3-10}
                        &                        & $\sigma= 0.035$    & $\sigma= 0.025$   & $\sigma= 0.015$    & $\sigma= 0.005$    & QF = 30    & QF = 50    & QF = 70    & QF = 90     \\ 
\hline
Baseline                & 23.65/\textbf{0.72}             & 20.98/0.51 & 21.55/0.58 & 22.28/\textbf{0.65} & 22.77/\textbf{0.68} & 20.36/0.58 & 21.01/0.60 & 22.01/\textbf{0.67} & 22.54/\textbf{0.70}  \\
Baseline + TCM                     & \textbf{29.24}/0.66             & \textbf{28.81}/\textbf{0.60} & \textbf{28.68}/\textbf{0.61} & \textbf{28.84/}0.61 & \textbf{28.85}/0.61 & \textbf{29.28}/\textbf{0.60} & \textbf{29.27}/0.60 & \textbf{28.75}/0.61 & \textbf{29.06}/0.58  \\
\hline
\end{tabular}
}
\end{table*}

\paragraph{Effectiveness of the TCM} TCM is an essential module for trajectory correction during the recovery process in the prompt-free framework. The effectiveness of this module is demonstrated by comparing Rows 1 and 3 in Table~\ref{tab:ablation}.The introduction of TCM substantially improves reconstruction fidelity, which improves PSNR from 22.75 dB to 34.65 dB, and SSIM increase from 0.74 to 0.97. These results indicate that TCM reduces the numerical errors accumulated and improves latent recovery accuracy.

To further compare the robustness of inversion strategies, text prompt-guided inversion and TCM are evaluated under different distortion. As shown in Table~\ref{tab:ablation_inver}, text-prompt-based inversion becomes less reliable under distortions. This is because text prompts cannot capture the latent deviation caused by image degradation, which limits their ability to stabilize the inversion trajectory. In constant, TCM addresses this limitation by correcting trajectories and stabilizing the reconstruction process to preserve high fidelity under noise.

\begin{table}[htbp]
\centering
\caption{Ablation on the number of affine coupling blocks.}
\label{tab:ablation_blocks}
\resizebox{1\linewidth}{!}{
\begin{tabular}{cccccc} 
\hline
\multirow{2}{*}{Number of CACM} & \multirow{2}{*}{NIQE$\downarrow$} & \multirow{2}{*}{$\mathrm{FID}_{(\text{ste},\text{real})}$$\downarrow$} & \multicolumn{2}{c}{Secret/ Recovery image pairs} & \multirow{2}{*}{Detection error(\%)}  \\ 
\cline{4-5}
                                &                       &                      & PSNR$\uparrow$  & SSIM $\uparrow$                                    &                                       \\ 
\hline
1                               & 4.86                     & 49.61                    & 30.49     & 0.97                                        & 0.56                                     \\
2                               & 4.18                     & 48.81                    & 31.02    & 0.97                                        & 0.50                                     \\
4                               & 3.07                  & 8.670               & 47.08 & 0.99                                     & 0.49                                  \\
6                               & 3.89                     & 12.11                    & 37.19     & 0.99                                        & 0.48                                     \\
\hline
\end{tabular}
}
\end{table}

\paragraph{Ablation on the Number of CACM Blocks}
We further evaluate the impact of using different numbers of CACM blocks, as reported in Table~\ref{tab:ablation}. The results reveal that the performance of the framework is sensitive to the choice of $K$. With a small value of $K$ (e.g., $K<4$), the mapping from the secret image to the latent space remains insufficient, and the reconstruction fidelity tends to be limited. Conversely, with an excessively large $K$, the overall performance tends to decrease. This degradation stems from the progressively accumulated information loss across successive invertible transformations.The system get the best overall performance with $K=4$. The above results suggest that an appropriate number of coupling blocks enables effective latent transformation and reduces excessive information loss, which leads to high visual quality of reconstruction.

\section{Conclusion}\label{conclusion}
This paper presents a prompt-free generative image steganography framework to improve the security and robustness of the steganography system. The proposed framework removes the dependence on transmitted text prompts and instead uses semantic priors extracted from reference datasets to enable user-controllable stego image generation. Within this framework, the hiding stage maps the secret image to the latent representation through CACM and uses SADM to guide controllable stego image generation with semantic priors. During recovery, the stego image is inverted to the latent representation through TCM, which uses single-step feedback to improve recovery stability under transmission distortions. Extensive experiments demonstrate that the proposed framework achieves competitive visual quality, stronger controllability, and improved robustness compared with existing methods. Future work will focus on increasing embedding capacity and enhancing recovery fidelity under complex real-world degradations.

\section*{Acknowledgments}
This research was supported by the National Natural Science Foundation of China (Grant Nos. 62376238, 62372170, 12571591), the Scientific Research Fund of Hunan Provincial Education Department (Grant No.2023JGSZ032), and the Postgraduate Scientific Research Innovation Project of Hunan Province (Grant No.CX20250999).

\bibliographystyle{IEEEtran}
\bibliography{refs}

\begin{IEEEbiography}[{\includegraphics[width=1in,height=1.25in,clip,keepaspectratio]{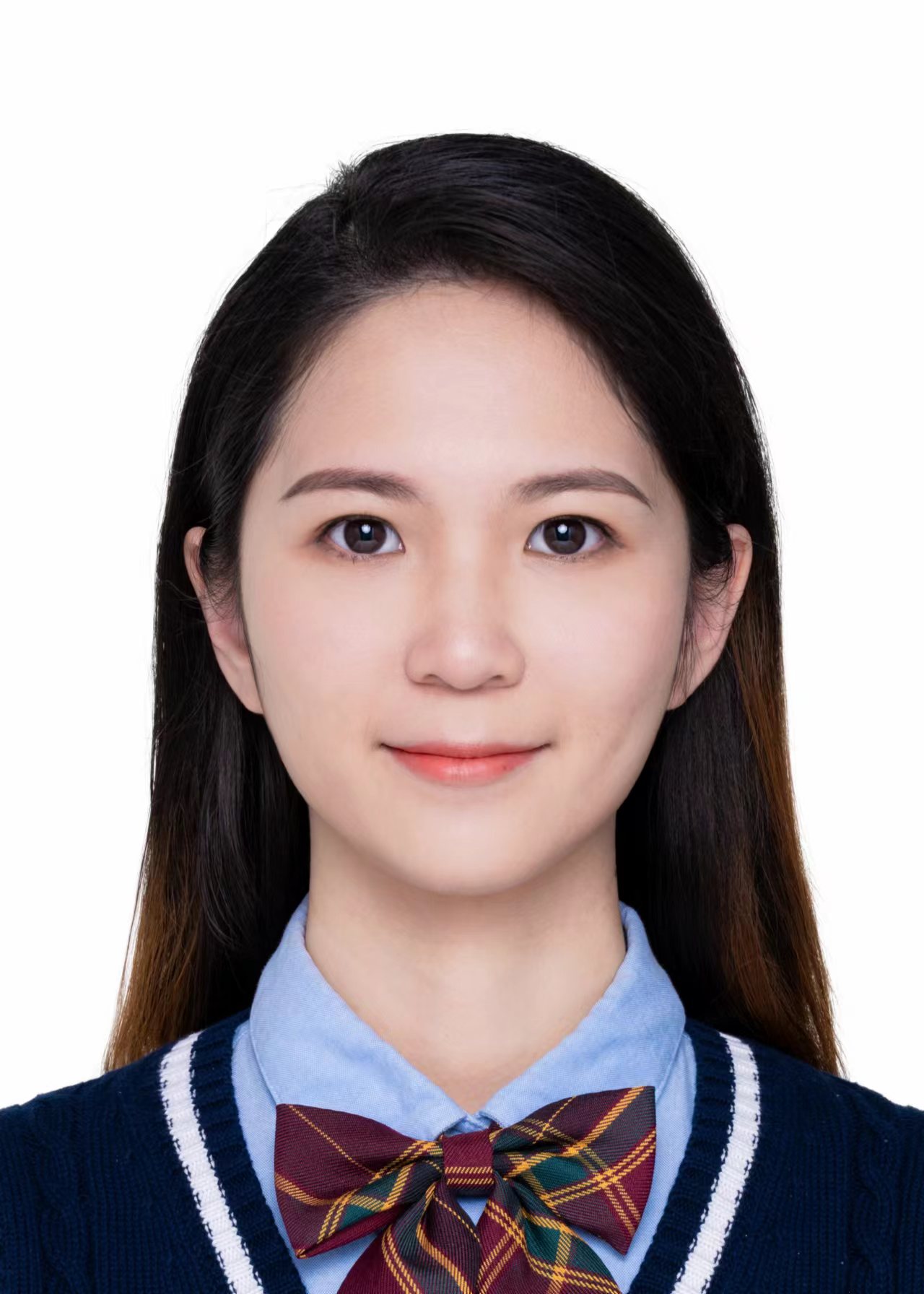}}]{Jingwen Cai}
received the M.E. degree in computer technology in 2022 from the Guilin University of Electronic Technology, Guilin, China, where she is currently working toward the Ph.D. degree in computer technology with the School of Computer Science, Xiangtan University, Xiangtan, China. Her research interests include information hiding, steganography, covert communication, and multimedia security.
\end{IEEEbiography}

\begin{IEEEbiography}[{\includegraphics[width=1in,height=1.25in,clip,keepaspectratio]{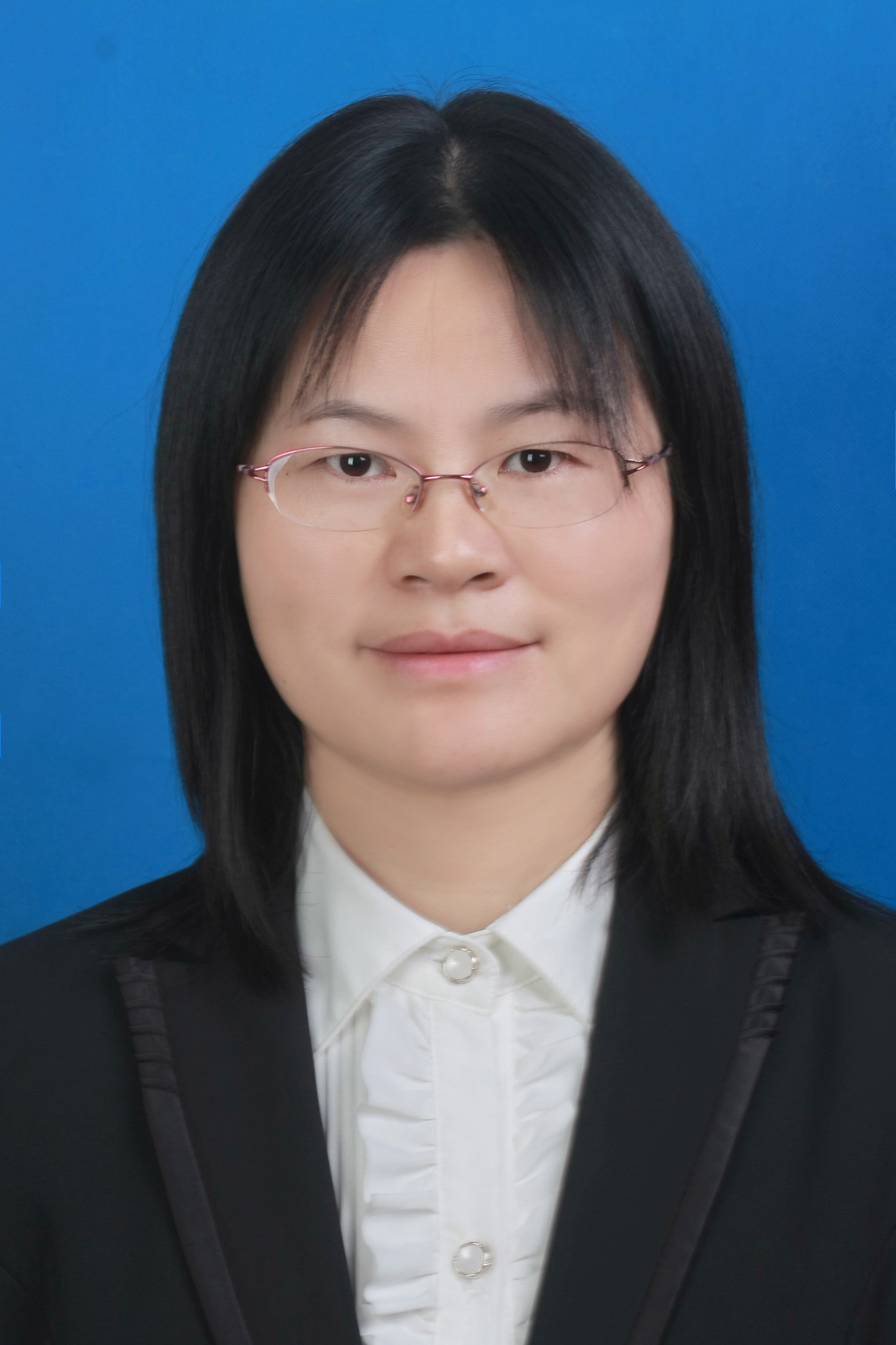}}]{Fen Xiao}
received the B.E. and Ph.D. degrees from Xiangtan University, Xiangtan,
China, in 2002 and 2008, respectively.
She was a Visiting Scholar at the State Key Laboratory of Pacific Northwest Pacific, Richland, WA, USA. She is currently a Professor with the School of Computer Science, School of Cyberspace Security, Xiangtan University. Her current research interests
include visual saliency detection, remote sensing, image analysis, and image description.
\end{IEEEbiography}

\begin{IEEEbiography}[{\includegraphics[width=0.95in,height=1.25in, clip]{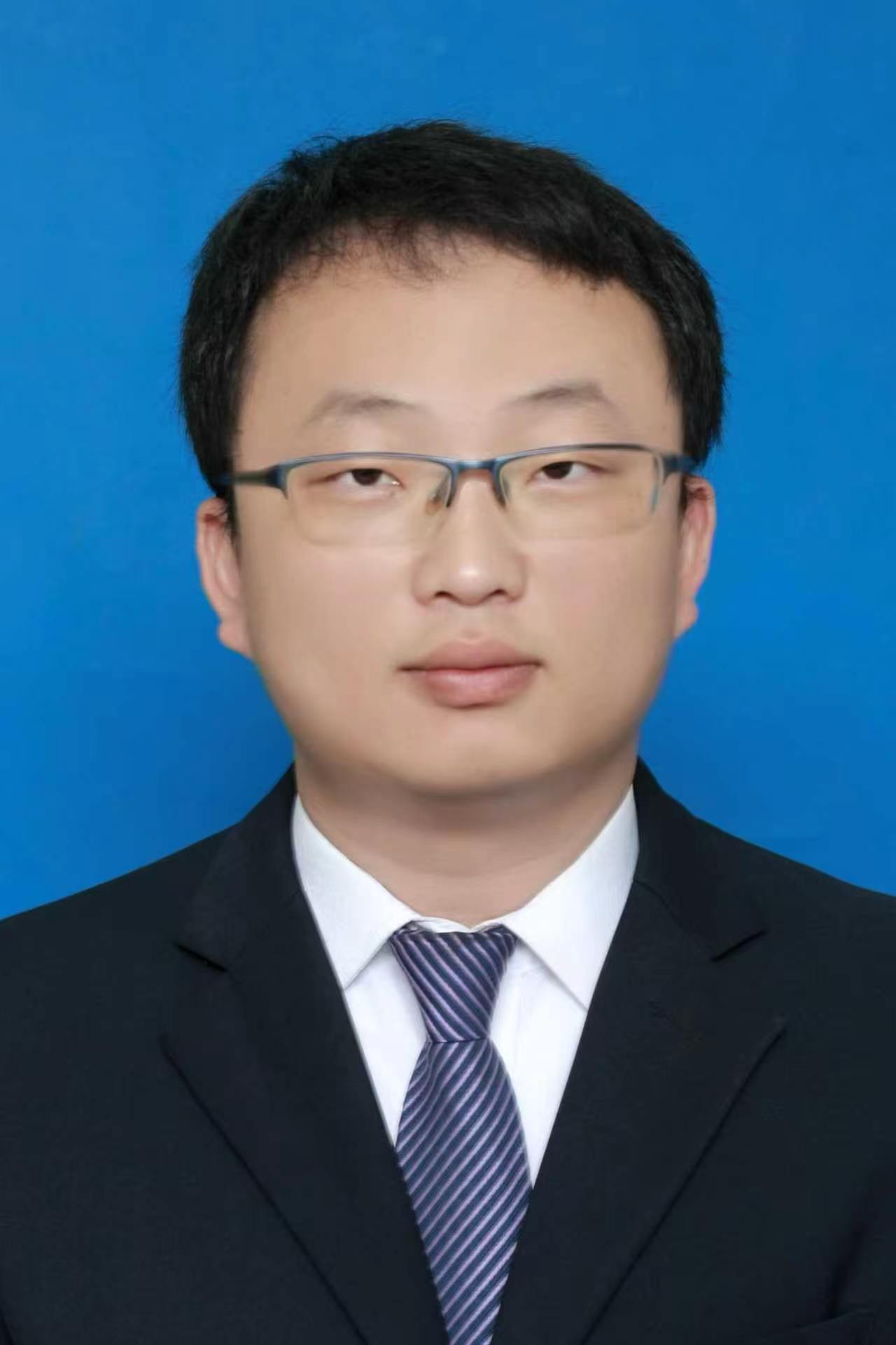}}]{Shuhua Deng} received the B.S. degree in computer science and the Ph.D. degree in computational mathematics from Xiangtan University, Hunan, China, in 2013 and 2018, respectively. He is currently an Associate Professor at the School of Computer Science, Xiangtan University, China. His current research interests include software-defined networks, network security, and machine learning. 
\end{IEEEbiography}

\begin{IEEEbiography}[{\includegraphics[width=1in,height=1.25in,clip,keepaspectratio]{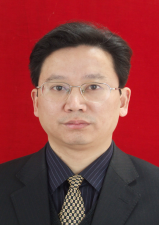}}]{Xieping Gao}
received the B.S. and M.S. degrees from Xiangtan University, Xiangtan, China, in 1985 and 1988, respectively, and the Ph.D. degree from Hunan University, Changsha, China, in 2003. He was a Visiting Scholar with the National Key Laboratory of Intelligent Technology and Systems, Tsinghua University, Beijing, China, from 1995 to 1996, and the School of Electrical and Electronic Engineering, Nanyang Technological University, Singapore, from 2002 to 2003. He is currently a Professor with the Hunan Provincial Key Laboratory of Intelligent Computing and Language Information Processing, Hunan Normal University, Changsha. His research interests include the areas of neural networks, evolution computation, and hyperspectral image processing.
Dr. Gao is a regular reviewer for several journals and he has been a member of the technical committees of several scientific conferences.
\end{IEEEbiography}
\end{document}